\documentclass[letterpaper,journal]{IEEEtran}
\usepackage{amsmath,amsfonts}
\usepackage{algorithmic}
\usepackage{algorithm}
\usepackage{array}
\usepackage[caption=false,font=normalsize,labelfont=sf,textfont=sf]{subfig}
\usepackage{booktabs,tabularx}
\usepackage{textcomp}
\usepackage{stfloats}
\usepackage{url}
\usepackage{verbatim}
\usepackage{graphicx}
\usepackage{cite}
\usepackage{makecell}
\usepackage{fontawesome5}
\usepackage{hyperref}
\usepackage[nolist]{acronym}
\usepackage{enumitem}
\usepackage{newtxtext,newtxmath}
\usepackage[table]{xcolor}
\setlist[enumerate]{topsep=.2cm}
\setlist[itemize]{topsep=.2cm}
\usepackage{placeins}

\newcolumntype{Y}{>{\raggedright\arraybackslash}X} 
\newcolumntype{L}[1]{>{\raggedright\arraybackslash\hspace{0pt}}p{#1}}

\newcommand\myparagraph[1]{\vspace{5pt}\noindent\textbf{#1}}
\newcommand\etal{\textit{et al.}}

\newcommand{\cautoref}[1]{%
  \begingroup
    \def\sectionautorefname{Section}%
    \def\subsectionautorefname{Subsection}%
    \def\subsubsectionautorefname{Subsubsection}%
    \def\figureautorefname{Figure}%
    \def\tableautorefname{Table}%
    \def\equationautorefname{Equation}%
    \def\appendixautorefname{Appendix}%
    \autoref{#1}%
  \endgroup
}

\usepackage{listings}
\begin{document}

\title{%
  LLMStructBench: Benchmarking Large Language Model Structured Data Extraction
}


\author{
    \IEEEauthorblockN{
        Sönke Tenckhoff\textsuperscript{*},
        Mario Koddenbrock,
        Erik Rodner\textsuperscript{*}
    }
    \thanks{
        *Corresponding authors, \\
        firstname.lastname@htw-berlin.de \\
        KI‑Werkstatt / FB2, University of Applied Sciences Berlin
    }
}





\maketitle

\begin{abstract}
We present \emph{LLMStructBench}, a novel benchmark for evaluating \acp{llm} on extracting structured data and generating valid \ac{json} outputs from natural-language text. 
Our open dataset comprises diverse, manually verified parsing scenarios of varying complexity and enables systematic testing across 22 models and five prompting strategies. 
We further introduce complementary performance metrics that capture both token-level accuracy and document-level validity, facilitating rigorous comparison of model, size, and prompting effects on parsing reliability.

In particular, we show that choosing the right prompting strategy is more important than standard attributes such as model size. This especially ensures structural validity for smaller or less reliable models but increase the number of semantic errors. Our benchmark suite is an step towards future research in the area of \acp{llm} applied to parsing or \ac{etl} applications.
\end{abstract}


\begin{acronym}
\acro{doc}[DOC]{Document Overall Correctness}
\acro{gt}[GT]{Ground Truth}
\acro{ie}[IE]{Information Extraction}
\acro{etl}[ETL]{Extract, Transform, Load}
\acro{json}[JSON]{JavaScript Object Notation}
\acro{llm}[LLM]{Large Language Model}
\acro{mk}[MK]{Missing Key}
\acro{mv}[MV]{Missing Value}
\acro{wv}[WV]{Wrong Value}
\end{acronym}

\section{Introduction}\label{sec:introduction}
\noindent
\IEEEPARstart{E}{xtracting} structured information from natural language is a frequent requirement in domains such as human resources~\cite{Palshikar_2023}, finance~\cite{Zheng_2019}, logistics, and e-commerce. \acp{llm} are increasingly used for this task or related \ac{etl} application, as for example highlighted by Xu~\etal~\cite{chen2024} in their survey on generative information extraction. However, their reliability in producing both syntactically valid and semantically correct outputs remains insufficiently understood. Lu~\etal~\cite{Yaxi_2025} demonstrate that even state-of-the-art models often fail to generate fully schema-compliant and semantically faithful \ac{json} outputs. Yet many downstream systems depend on receiving a complete and accurate object on the first attempt, making both structure and content essential~\cite{schick2023toolformer}.

To address this, we created a benchmark of realistic scenarios, each linked to a predefined \ac{json} schema and paired with natural language messages that contain all required information. 

Our dataset was synthetically generated with controlled prompting and then manually verified, removing or correcting cases where the text and schema did not align. It covers a range of complexity from flat key–value pairs to deeply nested structures. This design allows systematic testing of different models and prompting strategies while keeping \acp{gt} consistent and unambiguous. 

In our evaluation, we focus exclusively on open-source \acp{llm}. 
This choice reflects typical deployment constraints in data integration and \ac{etl} workflows, where sensitive or proprietary information cannot be shared with commercial cloud APIs. 
By benchmarking openly available models, we aim to provide realistic insights for practitioners who require locally deployable and privacy-preserving solutions. \\

For completeness, we additionally include the proprietary \textit{GPT-4o} model as a reference point.
Its inclusion allows us to gauge how far open-weight systems have progressed toward commercial performance levels, without changing our overall focus on openly available models.

The contributions of our paper can be summarized as follows:
\begin{enumerate}
\item We present an open dataset for structured parsing evaluation, comprising 995 manually verified samples across multiple use-case scenarios. It is specifically designed for JSON outputs and the application of \ac{llm}s (\cautoref{sec:dataset}).
\item We introduce quality metrics that allow for capturing syntactic and semantic errors both on a character-/ and the document-level (\cautoref{sec:metrics}).
\item In our work, we are able to compare 22 state-of-the-art \acp{llm} with different prompting strategies offering immediate results and recommendations for practitioners (\cautoref{sec:experiments} and \cautoref{sec:conclusions}).
\end{enumerate}

\section{Related Work}\label{sec:relatedwork}
\noindent
Recent studies explore two complementary directions for improving the reliability of structured outputs from \acp{llm}: 
(i) the development of benchmarks and decoding frameworks to enforce schema compliance~\cite{geng2025,Yaxi_2025} and,
(ii) the design of prompting strategies and workflows to improve extraction accuracy under realistic text conditions~\cite{li2024,brian2025,chen2024}. 

\myparagraph{Benchmarks and schema-constrained decoding.}
Geng~\etal~\cite{geng2025} introduce \textit{JSONSchemaBench}, a large benchmark of real-world \ac{json} Schemas that evaluate constrained decoding engines across efficiency, coverage of \ac{json} schema features, and output quality. Their results indicate that grammar- or schema-constrained decoding can accelerate generation and improve downstream task accuracy, yet support for advanced schema features differs substantially between frameworks. Our work is similar to their paper by measuring not only document validity, but also semantic extraction quality. In contrast to \cite{geng2025}, which focuses on engine compliance under given schema, we analyze how prompting strategies and model families trade off between perfect structure and correct values in an end-to-end \ac{ie} setting.

\myparagraph{Synthetic benchmark generation from structured data.}
\textit{StructText}~\cite{kashyap2025} proposes an automatic \textit{table-to-text} pipeline that uses existing tables as \ac{gt} to synthesize narratives and then evaluate text-to-table extraction along factuality, hallucination, coherence, and numeric or temporal accuracy. The authors find that models preserve numeric and temporal content, but often produce narratives that hinder machine extractability. Our dataset targets naturally occurring messages rather than \ac{llm}-synthesized narratives, which reduces the risk that coherence artifacts of synthetic text bias extraction results. At the same time, our analysis echoes their observation that accessibility for downstream parsing is as important as factuality.

\myparagraph{Domain benchmarks for structured spatial outputs.}
Li~\etal~\cite{yue2024} present a benchmark for GeoJSON generation and a hybrid R-tree-enhanced workflow that improves spatial coherence. While their findings generalize the need for structure-aware workflows, their scope is geospatial. Our benchmark complements this by focusing on general \ac{json} key–value extraction in heterogeneous messages and by reporting micro-averaged validity to reflect practical parser robustness. 

\myparagraph{Prompting strategies for structured \ac{ie}.}
Li~\etal~\cite{li2024} propose a two-step ``Generate \& Organize'' prompting scheme that first elicits free-form answers and then organizes them into the target structure with a task-specific clean-up stage. This approach reduces formatting errors that often occur when content extraction and formatting are requested at once. Our experiments corroborate the central premise: separating schema conformance from semantic correctness can change the error profile. 

\myparagraph{Structured outputs in competitive pipelines.}
Pradhan~\etal~\cite{brian2025} demonstrate that enforcing a fixed \ac{json} schema during generation, combined with lightweight domain-specific validation rules, can yield top leaderboard results without task-specific fine-tuning on several health \ac{ie} tasks. 
Their approach constrains the model to produce outputs that directly match the required structure. 
This concept, often referred to as \emph{schema-driven prompting}, provides the model with an explicit output schema or example object within the prompt. Our evaluation extends this idea to a broader range of open-source models and analyzes how such schema enforcement influences different error types, including \ac{mk}s and incorrect values.

\myparagraph{Surveys and practical extraction workflows.}
The survey by Xu~\etal~\cite{chen2024} synthesizes progress in generative \ac{ie} and highlights persistent challenges such as misalignment between natural language and structured targets, hallucination, and high computational cost. Our results quantify this misalignment for \ac{json} outputs by separating document validity from extraction accuracy.

\myparagraph{Positioning.}
Across these threads, prior work either stresses schema compliance under constrained decoding \cite{geng2025} or shows that careful prompting can make outputs parsable without training \cite{li2024,brian2025}. 
We contribute a complementary benchmark and analysis centered on \ac{json} parsing fidelity in everyday email communication, a common medium for automating administrative and business processes. 
Our evaluation quantifies how prompting strategies redistribute errors between structural validity and value correctness and introduces metrics that capture both dimensions. 
This addresses a gap identified by Kashyap~\etal~\cite{kashyap2025} and Xu~\etal~\cite{chen2024}, namely that factual accuracy alone is insufficient unless information remains accessible to automated parsers.

\section{Dataset and Benchmark}\label{sec:dataset}
\noindent
The proposed benchmark dataset is designed to evaluate \acp{llm} on structured \ac{ie} and \ac{json} generation tasks across diverse, realistic communication scenarios, something we will refer to as \emph{use-cases} in the following. These use-cases are inspired by typical email-based workflows in domains such as IT support, human resources, or project management—settings in which automation often relies on transforming free-text messages into structured, machine-readable records. Table~\ref{tab:use-case} provides a full list of all use-cases considered.

\begin{figure*}[!tb]
    \centering
    \includegraphics[width=1\linewidth]{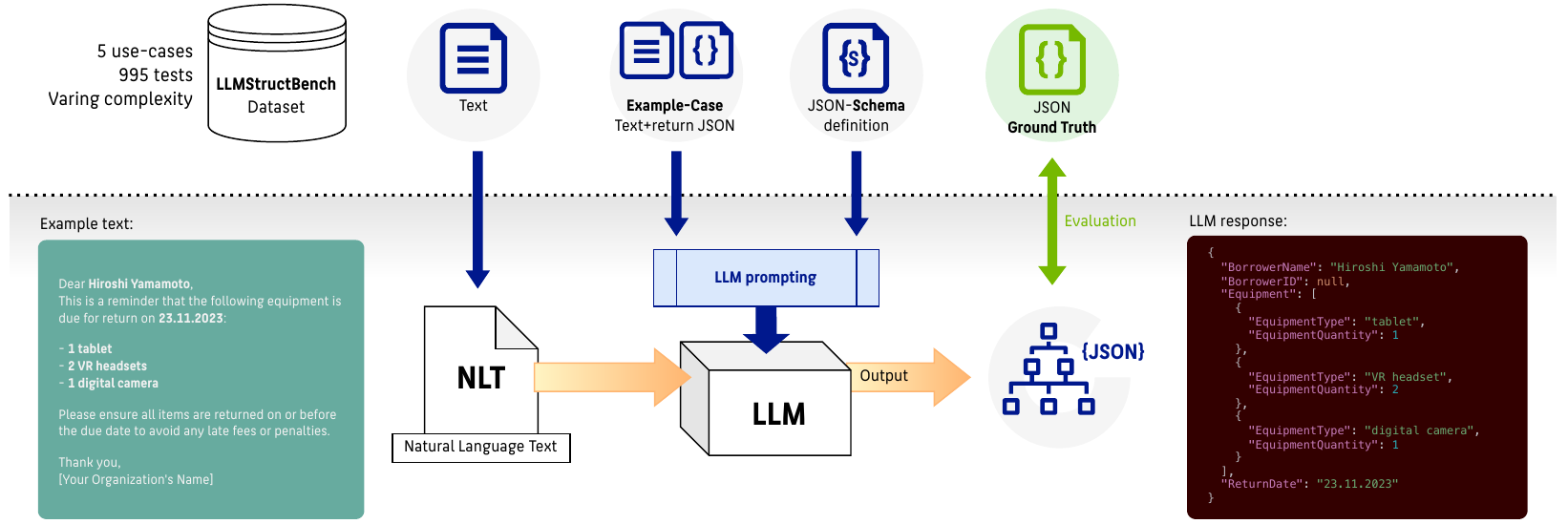}
    \caption{Schematic of the \emph{LLMStructBench} inference-time evaluation setup. Each test case provides a natural-language message, a corresponding \ac{json} schema, and an example input-output pair as input to the \ac{llm}. The model's generated \ac{json} object is then evaluated against the \ac{gt} for both syntactic validity and semantic accuracy.}
    \label{fig:dataset}
\end{figure*}

\myparagraph{Benchmark setup.}
\autoref{fig:dataset} illustrates the inference-time setup of our benchmark.  
Each model receives a natural-language message, a representative example pair (text and reference \ac{json}), and the corresponding \ac{json} schema that defines the required structure.  
It is instructed to return a single \ac{json} object adhering to this schema.  
The generated output is then compared to the validated \ac{gt} object to measure both structural validity and semantic accuracy. Further details on the evaluation procedure are provided in Section~\ref{sec:metrics}.

\myparagraph{Scenario and dataset structure.}
Each benchmark scenario (\emph{use case}) defines a self-contained extraction task and provides all necessary components for reproducible testing.  
A scenario specification includes:

\begin{itemize}
    \item a short description of the task
    \item the corresponding \ac{json} schema defining key names, data types, and nesting depth,
    \item example input (text) and output (\ac{json}) pairs 
    \item prompt templates for generating synthetic input text from structured data (\emph{email generation}) and for extracting structured data from text (\emph{conversion})
    \item a total of 199 validated test cases.
\end{itemize}

Each individual test case then links one \emph{natural-language message} with its corresponding schema-conform \emph{ground-truth object}.  
The schema remains fixed within a scenario, while the natural-language formulations vary to reflect realistic message diversity.

\myparagraph{Dataset composition.}
The complete dataset covers five use cases of varying structural complexity, resulting in 995 total tests.  
As shown in \autoref{tab:use-case}, the scenarios range from shallow key–value mappings (\textit{Support tickets}) to nested object arrays (\textit{Loan requests}), enabling analysis of model robustness across different structure depths. Two exemplary input-output pairs can be found in Appendix~\ref{app:dataset_examples}.

\begin{table}[htbp]
  \centering
  \setlength{\tabcolsep}{4pt}
  \caption{Overview of the \emph{LLMStructBench} dataset scenarios (use cases): Each scenario includes 199 unique, validated tests, detailed \ac{json} complexity specifications, and a brief description.}
  \label{tab:use-case}
  \footnotesize

  \begin{tabularx}{\linewidth}{L{1.5cm} Y c c c c}

    & & & \multicolumn{3}{c}{\ac{json}} \\
    \cmidrule(lr){4-6}
    Use case & Description & Count & Keys & Leaves & Depth \\
    \midrule

    Support tickets           & Processing of IT support ticket requests        & $199$ & $5$   & $5$  & $2$ \\
    \rowcolor{gray!10}
    Sick leave                & Processing of sick-leave requests               & $199$ & $7$   & $5$  & $3$ \\
    \rowcolor{white}
    Project extension         & Processing of project extension requests        & $199$ & $9$   & $6$  & $3$ \\
    \rowcolor{gray!10}
    Conference registration   & Processing of conference registration requests  & $199$ & $9$   & $7$  & $3$ \\
    \rowcolor{white}
    Loan request              & Processing of equipment loan requests           & $199$ & $10.04$* & $9.04$* & $4$ \\
    \midrule
    \multicolumn{6}{l}{\footnotesize * Average value; structure contains array of objects.} \\
    \bottomrule
  \end{tabularx}
\end{table}

\subsection{Synthetic Data Generation}
For each use case, synthetic test cases were generated using \textit{GPT-4o}. The process first produced fully populated \ac{json} objects that followed the scenario-specific schema. Using the email generation prompt, these \ac{json} objects were transformed into realistic and varying natural language messages, for example vacation requests, expense reports, or order confirmations. This ensured that the dataset covered diverse writing styles and lexical variations while maintaining a one-to-one correspondence between the \ac{gt} \ac{json} and the generated text.

To introduce variety, prompt instructions enforced randomization of entity names, dates, numerical values and other scenario-specific fields. This allowed the creation of many unique instances for each use case, covering both simple and highly nested data structures as displayed in \autoref{tab:use-case}.

\subsection{Dataset Verification and Cleaning}
The initial dataset was manually reviewed to ensure correctness. Faulty scenarios, where the \ac{gt} \ac{json} systematically deviated from the information present in the corresponding text, were removed entirely. In cases where only specific entries were affected, for example a single missing value repeated across multiple instances, corrections were applied by hand.

After these steps, all retained examples contain:
\begin{itemize}
    \item a complete and schema-valid \ac{json} object
    \item a corresponding text message reflecting all information present in the JSON object
    \item no systematic mismatches between the two representations
\end{itemize}

\subsection{Final Dataset Composition}
The cleaned dataset covers multiple complexity levels, from flat key–value structures to deeper nested hierarchies with lists and objects. Scenarios include realistic domains such as human resources, project management, IT support, logistics and e-commerce. This diversity allows the benchmark to assess both the syntactic reliability and the semantic accuracy of model outputs under controlled and realistic conditions. 
We also experimented with use cases and schema that involved more nested \ac{json}-objects, however, this did not result in even medium quality generated examples and have been therefore skipped. Often times the generation of the email was missing out on many crucial values from the ground-truth base object.

\section{Quality Metrics and Evaluation Strategy}\label{sec:metrics}
\noindent 
To evaluate outputs against the \ac{gt}, we perform a recursive comparison and classify discrepancies into:
\begin{itemize}
    \item \textbf{\acf{mk}:} Key–value pairs present in the reference but absent in the model output.
    \item \textbf{\acf{mv}:} Values that are absent, either because their key is missing entirely or because the key is present but its value is \texttt{null}.
    \item \textbf{\acf{wv}:} Values that differ, subdivided into:
    \begin{itemize}
      \item \emph{Value deviation:} Both reference and candidate are strings; measured by Levenshtein distance.
      \item \emph{Value error:} Same primitive type but different content.
      \item \emph{Coercible mismatch:} Stringified forms match, but the underlying types differ.
      \item \emph{Type error:} Neither same type nor coercible to the same string.
    \end{itemize}
\end{itemize}

\subsection{Performance Metrics}\label{sec:promp-performance}

\myparagraph{Micro-averaged F1 ($F1_{\text{micro}}$).}\label{sec:metric-f1}
We treat each \texttt{key:value} pair as a classification instance and measure precision/recall over \emph{all} instances in the corpus.  
Because downstream consumers usually care more about the correctness of the produced values than about the mere presence of a key, we split the metric into two parts and mix them linearly
\begin{align}
  F1_{\text{micro}} &= \alpha\,F1_{\text{keys}} + (1-\alpha)\,F1_{\text{values}}
  \label{eq:f1}
\end{align}
with a \textbf{key weight} of $\alpha = 0.25$ (ref. \ref{sec:weight-choice}). Concretely:

\begin{itemize}
  \item \textbf{Key component} ($F1_{\text{keys}}$) counts true-positive keys (present and unique), false-positives (extraneous keys) and false-negatives (\ac{mk}s).
  \item \textbf{Value component} ($F1_{\text{values}}$) evaluates value correctness with a nuanced credit system. It rewards exact matches fully and penalizes wrong or missing values. For mismatches, it distinguishes between error types:
    \begin{itemize}[leftmargin=*,topsep=0pt,itemsep=0pt]
        \item It grants partial credit for \emph{coercible mismatches} (e.g., string \texttt{"42"} vs. integer $42$) and for \emph{value deviations} in strings, where credit is proportional to string similarity (inverse Levenshtein distance).
        \item It actively penalizes string comparisons that are drastically different, discouraging nonsensical outputs.
        \item It assigns zero credit for hard \emph{type errors} and \emph{value errors}, treating them as complete failures.
    \end{itemize}
\end{itemize}
A formal definition of the full scoring function, including fuzzy value matching and weighting constants, is provided in Appendix~\ref{app:microf1_formula}. In essence, the metric rewards fully correct pairs, grants graded credit for close or type-compatible values, and penalizes missing or nonsensical entries, yielding a single bounded score $F1_{\text{micro}}\!\in\![0,1]$.

\myparagraph{Document success (\ac{doc}$_{\text{micro}}$).}\label{sec:metric-doc}
Token-level scores can still be high when every document has a few small errors, yet many real-world pipelines accept \emph{only fully valid JSON}.  
Therefore we add a document success term
\begin{align}
  DOC_{\text{micro}} \;=\;
  \frac{\text{\,\#\,Correct\,}}{\text{\,\#\,Tests\,}}
  \Bigl(1 - \frac{\text{\,\#\,Failed\,}}{\text{\,\#\,Tests\,}}\Bigr),
  \label{eq:doc}
\end{align}

i.e.\ the share of completely correct generations \emph{tempered} by the share of outright failures (empty output, unparsable \ac{json}, fatal schema violation). It, too, lies in~$[0,1]$. This metric corresponds to document‑level validity or schema adherence in structured output benchmarks \cite{geng2025}. It captures brittleness that token‑level scores can hide. \\

\subsection{Combined Rating}\label{sec:metric-overall-rating}

To rank models on a single, easily interpretable scale we combine \emph{token-level} accuracy with \emph{document-level} validity into one composite score, which we derive from $F1_{\text{micro}}$ and \ac{doc}$_{\text{micro}}$ scores.
\begin{align}
  \text{Score} &= (1-\lambda)\,F1_{\text{micro}} \;+\; \lambda\,DOC_{\text{micro}}
  \label{eq:score}
\end{align}
with a \textbf{weight} of $\lambda = 0.5$, as motivated in the paragraph below. 

Document‑level validity matches the notion of validity used by \textit{JSONSchemaBench} for schema adherence under constrained decoding \cite{geng2025}. Micro‑averaged extraction F1 follows the practice in \ac{ie} surveys and shared tasks where per field correctness is aggregated over all instances \cite{chen2024,brian2025}. \\

\myparagraph{Choice of weights.}\label{sec:weight-choice}
We selected the two free hyper-parameters empirically, guided by practitioner feedback:
\begin{enumerate}
  \item \textbf{Key to Value weight $\alpha = 0.25$.}  
        A low weight on key-level F1 reflects the fact that getting a value correct is typically more critical than getting its key correct. Furthermore, a \ac{mk} is implicitly penalized twice: once in the key-F1 score (as a false negative) and again in the value-F1 score (since its value is also missing). By down-weighting the key component, we heavily prioritize models that produce correct values in the correct locations, as a wrong value can silently corrupt downstream logic, whereas an extraneous or \ac{mk} is often less severe.
  \item \textbf{Document to F1 weight $\lambda = 0.5$.}  
        In early experiments ($0.3 \leq \lambda \leq 0.7$) we observed that scores below~$0.4$ de-emphasised outright failures (models that emit markdown instead of \ac{json} still ranked mid-table) while scores above~$0.6$ masked the nuanced differences between good and excellent token-level accuracy.  
        A neutral split (\emph{half structure, half content}) preserves sensitivity in both regimes.
\end{enumerate}

Because each term is differentiable and independently reported, practitioners can still drill down into \emph{why} a model under-performs. Whether due to structural brittleness (low~\ac{doc}$_{\text{micro}}$) or semantic drift (low~$F1_{\text{micro}}$).

\section{Model and Prompting Techniques}\label{sec:prompting-strategies}
\noindent
\subsection{Models Used in this Study}

We selected twenty-two model variants from five families, spanning from $0.6$ to $70$ billion parameters:
\begin{itemize}
  \item \textbf{Qwen3 \cite{bai2023qwen}:} Qwen3:0.6b (0.6\,B), Qwen3:1.7b (1.7\,B), Qwen3:4b (4\,B), Qwen3:8b (8\,B), Qwen3:14b (14\,B), Qwen3:32b (32\,B).
  \item \textbf{Phi \cite{abdin2024phi}:} Phi3:3.8b (3.8\,B), Phi3:14b (14\,B), Phi3.5:3.8b (3.8\,B), Phi4-mini:3.8b (3.8\,B), Phi4:14b (14\,B).
  \item \textbf{Gemma3 \cite{team2025gemma}:} gemma3:1b (1\,B), gemma3:12b (12\,B), gemma3:27b (27\,B).
  \item \textbf{Deepseek-R1 \cite{liu2024deepseek}:} Deepseek-R1:1.5b (1.5\,B), Deepseek-R1:7b (7\,B), Deepseek-R1:8b (8\,B), Deepseek-R1:14b (14\,B), Deepseek-R1:32b (32\,B), Deepseek-R1:70b (70\,B).
  \item \textbf{LLama3 \cite{grattafiori2024llama}:} Llama3.1:70b (70\,B), Llama3.3:70b (70\,B).
\end{itemize}

In addition, we include \textit{GPT-4o} as a closed-weight reference model.
It is evaluated under the same benchmark conditions to contextualize the performance of open models.
Since its parameter count and detailed training data remain undisclosed, it is excluded from size-based and combined analyses. The model evaluation serves solely as an upper-bound reference rather than part of the open-source comparison pool.

Besides that, the selection of open sourc models allows us to assess how both architecture and scale affect \ac{json} parsing performance and data quality.  

\subsection{Prompting Techniques}

To identify the most reliable way to get well-formed \ac{json} from our models, we evaluated five different prompting configurations (see \autoref{tab:json_comparison}). Each configuration varies along two axes: 
\begin{enumerate}
    \item whether we set the Ollama API format parameter (either to the literal string "json" or to a full \ac{json}-Schema object) and
    \item whether we explicitly instruct the model, via system and user prompts, to follow a provided schema and an example \ac{json} structure
\end{enumerate}

\begin{table*}[hbp]
\centering
\caption{Overview of the five \ac{json} prompting strategies evaluated in this study. Strategies vary based on the explicit `format` parameter setting in the API request and the inclusion of schema/example object guidance in the system prompt.} 
\label{tab:json_comparison}
\begin{tabular}{c L{5.9cm} L{5cm} L{5cm}} 
\toprule
\textbf{Tag} & \textbf{Description} & \makecell[l]{\textbf{"format" parameter}\\\textbf{(API-\ac{json} mode)}} & \textbf{System-prompt} \\
\midrule
\textbf{J}   & Only set the "format" parameter during the request to "json" & \faExclamationTriangle\ \texttt{"json"} (str) as request parameter & \faTimes\ No mentioning about structure \\
\addlinespace 
\textbf{P}   & Demand \ac{json} return in prompts with Schema and example object & \faTimes\ Not set & \faCheckCircle\ Include \ac{json} schema and example object \\
\addlinespace
\textbf{PJ}  & Set "format" parameter to "json" and prompt for json in prompt & \faExclamationTriangle\ \texttt{"json"} (str) as request parameter & \faCheckCircle\ Include \ac{json} schema and example object \\
\addlinespace
\textbf{J+}  & Set Schema into "format" parameter during request & \faCheckCircle\ \texttt{\ac{json}-Schema} (obj) as request parameter & \faTimes\ No mentioning about structure \\
\addlinespace
\textbf{PJ+} & \makecell[l]{Set "format" parameter and prompt\\ with Schema and example object} & \faCheckCircle\ \texttt{\ac{json}-Schema} (obj) as request parameter & \makecell[l]{\faCheckCircle\ Include \ac{json} schema and\\ example \ac{json} object} \\
\bottomrule
\end{tabular}
\end{table*}

By comparing these five conditions, we can disentangle the impact of API-level \ac{json} enforcement from explicit prompt-level schema guidance on the consistency and validity of the returned \ac{json}. 

\subsection{Prompting Technique Evaluation}

To determine which prompting strategy yields the most reliable \ac{json} generation across models, we computed a premature performance score for each strategy by weighting outcomes from the evaluation results. Specifically, each strategy was scored using a simple linear metric that prioritizes correctly parsed outputs and penalizes various types of errors with different severity:

\begin{itemize}
    \item \textit{Correct:} Flawless generations are rewarded with \\ \textbf{10 Points} per test-case. This is rewarded with a large amount of points, as we really take emphasis on consistency in data extraction. 
    \item \textit{Mistakes:} Value deviation or extra keys are penalized with \\ \textbf{-1 Point} per occurrence. Occurrences of mistakes enforce the need for data validation through other means in the downstream pipeline.
    \item \textit{Errors:} Missing values or keys are penalized with \\ \textbf{-2 Points} per occurrence. They cost double the amount as \textit{mistakes}, as they are really missing information, that need substitution later.
    \item \textit{Failures:} A failed generation is penalized with \\ \textbf{-15 Points} per test-case. Generally all scenarios and tests of the benchmark have a clear solution, so failing to return a JSON-Object in the first place is penalized servilely.
\end{itemize}

This weighted scoring scheme reflects the qualitative differences between error types and their impact on usability in downstream applications. For each model, the strategy with the highest score was selected as the "winner" for that model. In cases of a tie, all top-scoring strategies were counted. This procedure allowed us to tally how often each prompting configuration produced the most favorable result across the full model pool. It also allows for discovering shortcomings of individual models or entire model families, regarding specific prompting strategies.

\section{Experiments}\label{sec:experiments}
\noindent
We structure our evaluation into three stages.  
First, we compare all prompting strategies using the parsing metric introduced in Section~\ref{sec:prompting-parsing-eval}, and rank them in Table~\ref{tab:method-selection}.  

Second, in Section~\ref{sec:model-size-analysis}, we study the two best strategies across model families and sizes, evaluating both extraction accuracy and document validity via $F1_{\text{micro}}$ and \ac{doc}$_{\text{micro}}$ (Section~\ref{sec:metric-overall-rating}).  

Finally, Section~\ref{sec:leaderboard-results} presents a consolidated leaderboard of all evaluated models, shown in Table~\ref{tab:model-leaderboard}.

\subsection{Impact of the Prompting Strategy on Parsing performance}\label{sec:prompting-parsing-eval}

\begin{figure}[!tb]
    \centering
    \includegraphics[width=1\linewidth]{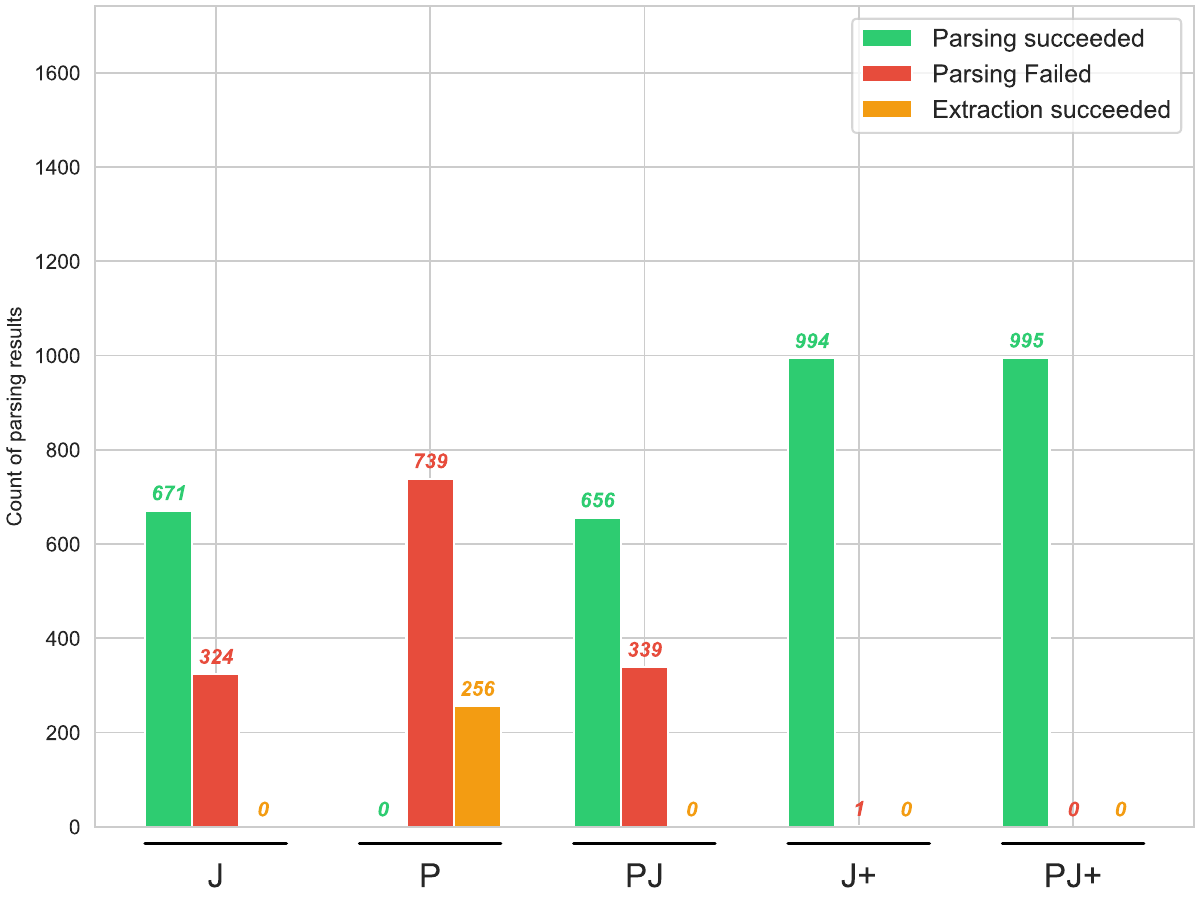}
    \caption{Parsing success rates for the \textit{Phi3 - 3.8b} model across different prompting strategies. Bars represent the count of test cases resulting in a perfectly parsed \ac{json} (green), successfully extracted \ac{json} from surrounding text (yellow), or a complete parsing failure (red).}
    \label{fig:prompt-parsing}
\end{figure}

We begin with one of the lower-ranked models, which struggled to produce a sufficient number of flawless \ac{json} results even under the more structured prompting setups. In \autoref{fig:prompt-parsing} you can see the evaluation of the \textit{Phi3} model with \textit{3.8b} parameters. The results are separated to display the capability of the selected model, to return a valid json object for each strategy of prompting. In this first evaluation, we don't take into account the quality of content of the answer itself, but rather focus only on the possibility of parsing the answer to a valid \ac{json}-Object. \\

The model displayed (\textit{Phi3 - 3.8b}), together with its successor (\textit{Phi3.5 - 3.8b}), is among the worst performing models within this category. Across all prompting strategies, there are at least some occasions where tests fail completely. This is visualized by the red bars on the chart.
On the other hand, the green bar represents all test counts where the model responded with a clean answer, that was successfully parsed to a \ac{json}-Object without any additional tweaks. \\

A count of \textit{Extracted parsing succeeded} is added every time the model returns more context around the wanted \ac{json}-Object. Here, we try to use regex to extract the relevant section, containing the \ac{json}-formatted string from the larger text-answer. \\

\begin{figure}[!tb]
    \centering
    \includegraphics[width=1\linewidth]{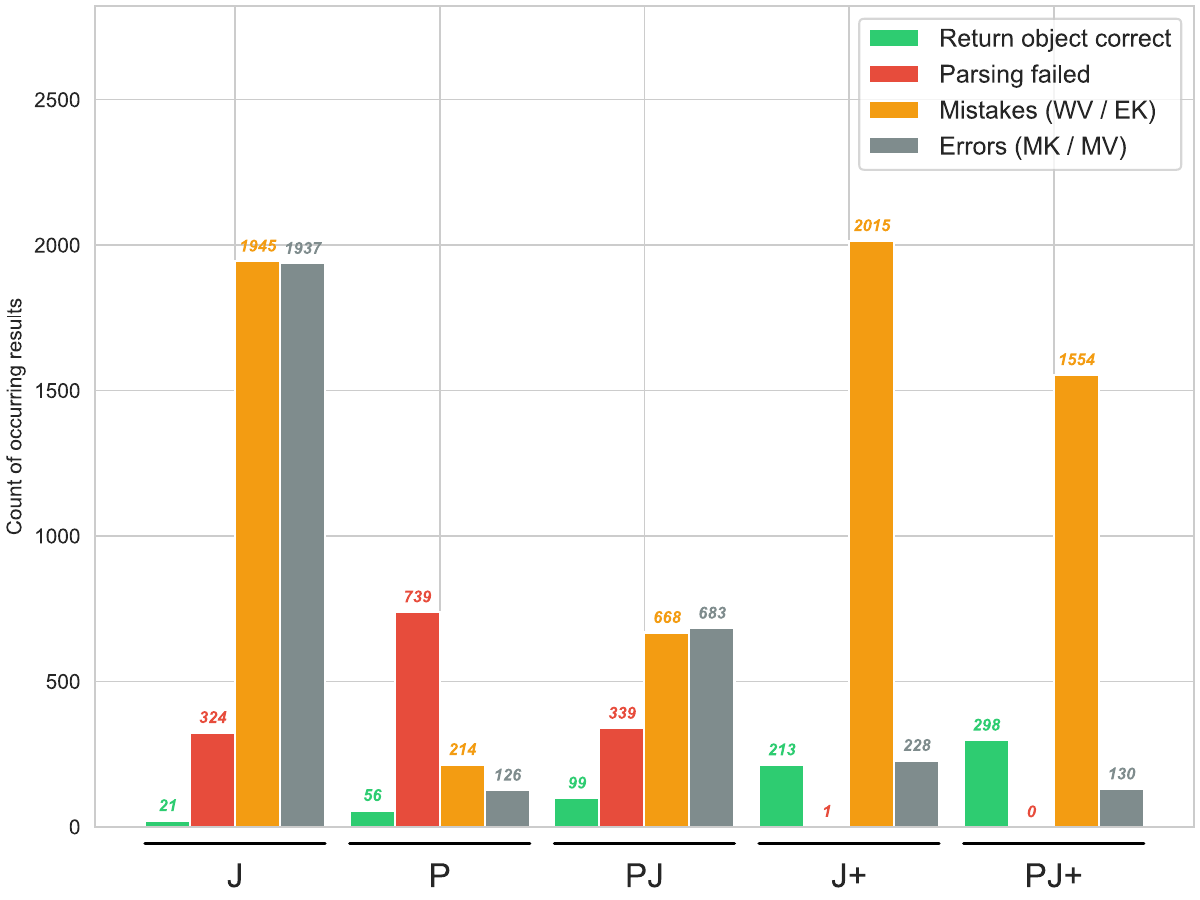}
    \caption{Distribution of error types for the \textit{Phi3 - 3.8b} model's \ac{json} outputs, categorized by prompting strategy. Categories include perfect matches to \ac{gt} (green), semantic mistakes (orange), critical errors (gray), and complete parsing failures (red).}
    \label{fig:prompt-error}
\end{figure}

To evaluate a model's capability to also extract the right information from a given text, again, compared by different prompting strategies, we have to look at another chart. On \autoref{fig:prompt-error} we compare not only parsing success, but this time how many test cases had a perfect match of the json return object to \ac{gt}. This count is destincly low for all prompting-strategies with the \textit{Phi3 - 3.8b} model.

We also notice quite large bars for \textit{mistakes} (orange) which indicates mismatches between \ac{json}-values or additional key/value pairs. These are not as severe as the \textit{errors} (gray) which indicate when a key, or value is completely missing from the language model's response. \\

When comparing the results based on prompting techniques for \textit{Phi3}, it demonstrated the most promising performance applying the \textbf{PJ+}-prompting strategy. While it produced $298$ perfect \ac{json} objects and incurred $130$ errors, its notably absence of failed test cases demonstrates the baseline of our prompting technique selection process. \\

After evaluating prompting techniques for all chosen models, we end up with a table (\autoref{tab:technique-eval}) of most promising prompting techniques for further testing and evaluation.

\begin{table}[tbp]
\centering
\caption{Aggregated performance across all models for each prompting strategy, categorized by outcome. Metrics include total counts for perfectly correct \ac{json} generations, semantic mistakes, critical errors (\ac{mk}/\ac{mv}), and complete parsing failures.}
\label{tab:technique-eval}
\begin{tabular*}{\linewidth}{@{\extracolsep{\fill}}c rrrr} 
\toprule
\textbf{Strategy} & \textbf{Correct} \faArrowUp & \textbf{Mistakes} \faArrowDown & \textbf{Errors} \faArrowDown & \textbf{Failed} \faArrowDown \\
\midrule
J   & $2,271$ & $45,951$ & $28,797$ & $726$ \\
\addlinespace 
\textbf{P}   &  \hspace{0.1mm} $\mathbf{7,791}$ &  \hspace{0.1mm} $\mathbf{14,171}$ & $2,879$ & $1,597$ \\
\addlinespace 
PJ   & $6,800$ & $15,792$ & $13,506$ & $445$ \\
\addlinespace 
J+   & $6,129$ & $28,903$ & $1,242$ & $135$ \\
\addlinespace 
\textbf{PJ+}   & $7,646$ & $20,652$ & \hspace{0.1mm} $\mathbf{1,169}$ & \hspace{0.1mm} $\mathbf{18}$\\
\addlinespace 
\bottomrule
\end{tabular*}
\end{table}

Based on this result and our proposed metric for scoring the different performances based on prompting strategy, we are able to deviate following strategy ranking (\autoref{tab:method-selection}). It clearly shows \textbf{P}-strategy as best performing one with a count of $11$ favorable model applications, with \textbf{PJ+} as runner-up.

This is also reflected in the raw evaluation data in \autoref{tab:technique-eval}, as both \textbf{P} and \textbf{PJ+} have the highest count of entirely \textit{correct} results. Prompting with \textbf{P} strategy exhibits an unusual high count of \textit{failed} test cases. This is down to the \textit{Phi3}-family really struggling with this specific prompting type. All other model families display a lower average count of failures using the \textbf{P}-prompting technique. 
While the \textbf{P} strategy almost shows double the amount of \textit{errors}, compared to \textbf{PJ+}, it's still in the lower range compared to other strategies. Additionally it is the best option for less severe \textit{mistakes} with $14.171$ combined occurrences over $995$ performed tests.

\begin{table}[tbp]
\centering
\caption{Ranking of prompting strategies based on their overall effectiveness, showing the count of models for which each strategy achieved the highest performance score.}
\label{tab:method-selection}
\begin{tabular*}{\linewidth}{@{\extracolsep{\fill}}l ccccc} 
\midrule
Strategy & \textit{J} & \textbf{P} & \textit{PJ} & \textit{J+} & \textbf{PJ+} \\
\midrule
Count & $0$ & \textbf{11} & $3$ & $0$ & \textbf{8} \\
\bottomrule
\end{tabular*}
\end{table}

Going forward, we are focusing on comparing results for both strategies proven to be relevant.

\subsection{Impact of Model Size on Value Extraction Performance}\label{sec:model-size-analysis}

\begin{figure}[!tb]
    \centering
    \includegraphics[width=1\linewidth]{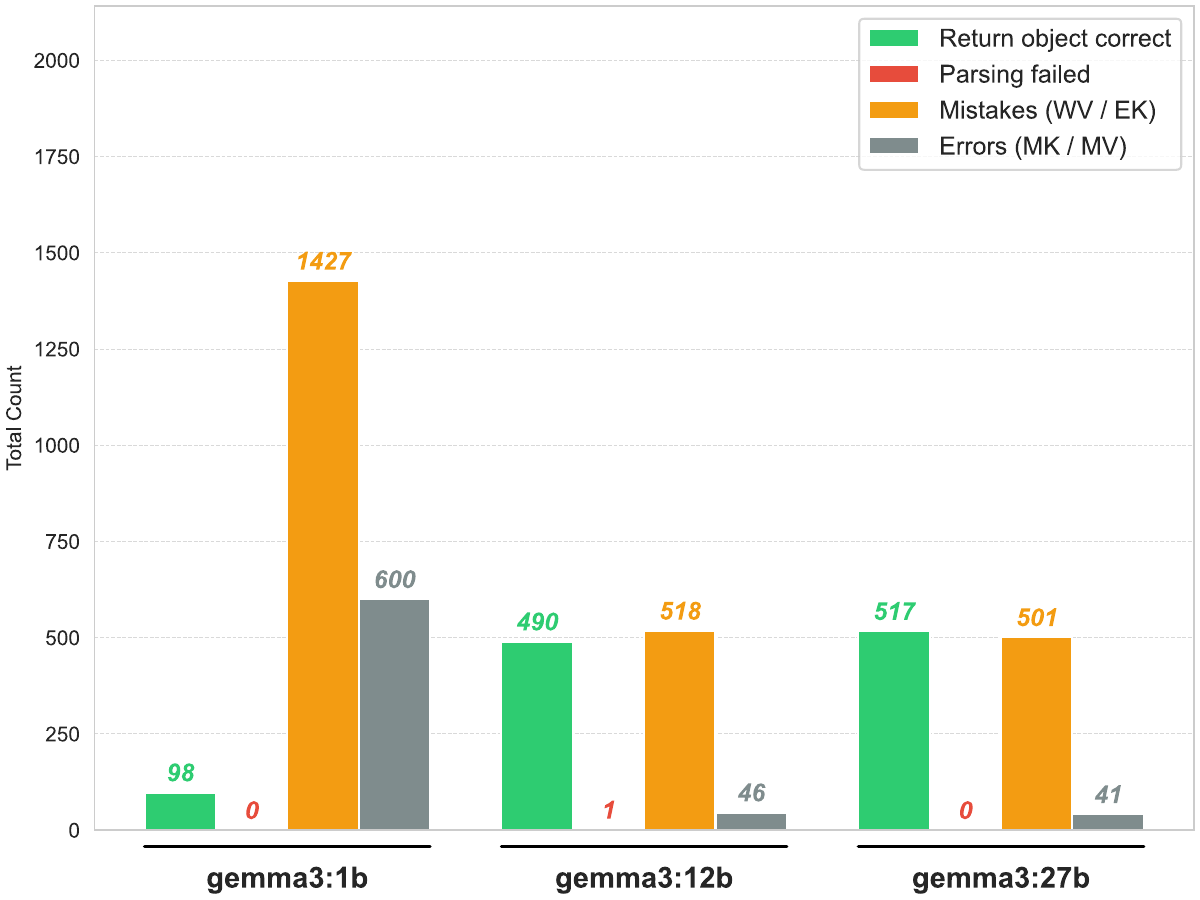}
    \caption{Performance breakdown by model size for the \textit{Gemma3} family using the \textbf{P}-prompting strategy. The stacked bars illustrate the counts of perfectly correct (green), mistaken (orange), erroneous (gray), and failed (red) \ac{json} extractions across model variants.}
    \label{fig:model_size_performance_p}
\end{figure}

Now, that we have established the prominent prompting strategies \textbf{P} and \textbf{PJ+} we can take a look at the different parsing performances by model family and respective size variation. In \autoref{fig:model_size_performance_p} we can observe the performance of all \textit{Gemma3}-model size variations. The graph clearly displays the steep improvement from a really small model size (e.g. $1B$ parameters) to the next bigger variant ($12B$ parameters). This is expressed by the reduction of \textit{errors} and \textit{mistakes}, at the same time as the response count of perfect matches with the \ac{gt} \ac{json} increases. \\

When we compare the second biggest, to the largest model variant, we assert only a small improvement in the increase of \textit{success} counts as well as the reduction of \textit{failure}, \textit{errors} and \textit{mistakes}. This is to be expected, as the model size only doubles from the second to the last model variant, while the second to the first is an approximated $12$-times increase in size.

While the comparison of model performance by the current four metrics, \textit{success}, \textit{fail}, \textit{error} and \textit{mistake} is reasonable for settling on a fitting prompting strategy, it isn't very well suited for the evaluation of model families including higher counts of model variants. \\

\begin{figure}[!tb]
    \centering
    \includegraphics[width=1\linewidth]{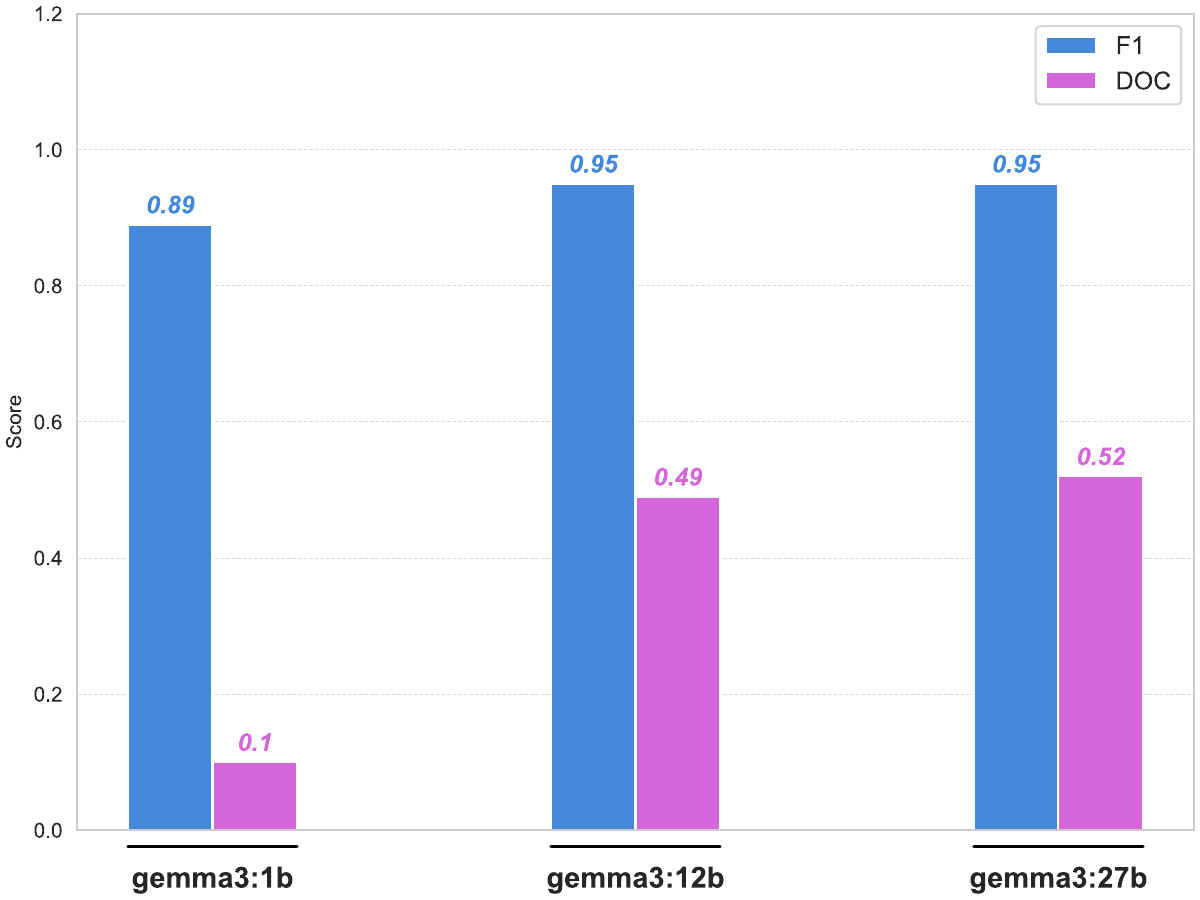}
    \caption{Composite scores ($F1_{\text{micro}}$ and $DOC_{\text{micro}}$) for the \textit{Gemma3} model family, demonstrating performance trends across different model sizes under the \textbf{P}-prompting strategy.}
    \label{fig:model_size_score_p}
\end{figure}

Therefore we apply the \emph{overall performance rating} that we proposed in a previous section (ref. Section~\ref{sec:metric-overall-rating}). It gives us one single score for each model variant, that is derived from the accuracy of extracting correct data ($F1_{\text{micro}}$-Score ref. Section~\ref{sec:metric-f1}), as well from the models ability to return parse-able and flawless \ac{json} objects in the first place (\ac{doc}$_{\text{micro}}$-Score ref. Section~\ref{sec:metric-doc}). For the \textit{Gemma3} family the overall scoring is shown in \autoref{fig:model_size_score_p}, still separated into $F1_{\text{micro}}$ and \ac{doc}$_{\text{micro}}$. \\

We can clearly observe the relation between the low count of flawless results ($98$ out of $955$) of the $1$b parameter variant in \autoref{fig:model_size_performance_p} to the bad performing \ac{doc}$_{\text{micro}}$-score ($0.1$) in \autoref{fig:model_size_score_p}. A perfect \ac{doc}$_{\text{micro}}$-Score can only be achieved when \emph{all} \ac{json}-Objects returned without any deviations to the \ac{gt}. Another observation between the error and score plots is the reduction of \textit{errors} and \textit{mistakes} with increasing model size to a slightly improving $F1_{\text{micro}}$-score. This micro averaged composite score also takes partially correct results, like string overlaps through Levenshtein-distance, or variable type deviations into account, thus dampening the impact of change in the raw count of error occurrences over model size.

\begin{figure}[!tb]
    \centering
    \includegraphics[width=1\linewidth]{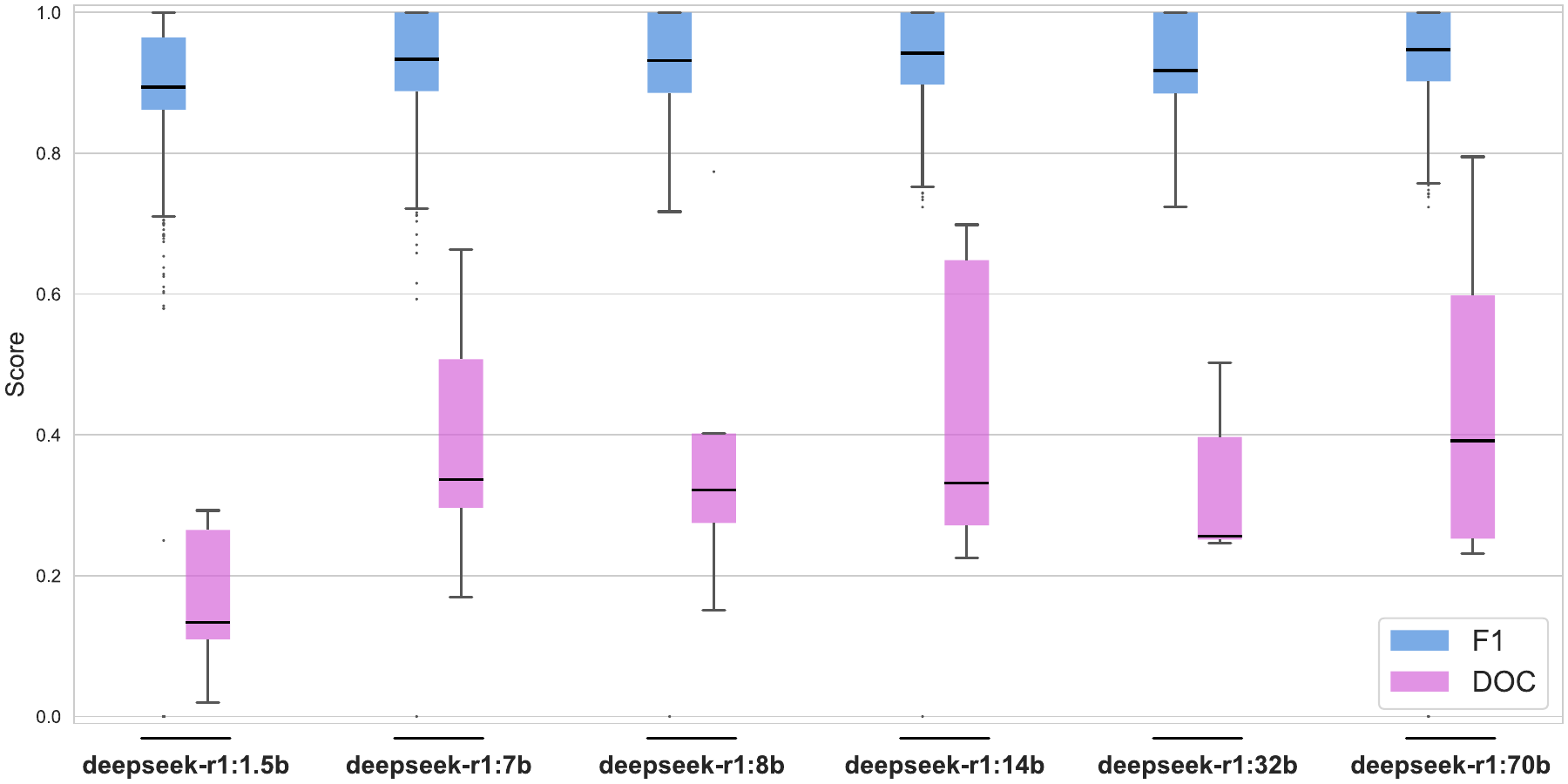}
    \caption{Distribution of $F1_{\text{micro}}$ (blue boxes) and $DOC_{\text{micro}}$ (purple boxes) scores for the \textit{Deepseek-R1} model family, illustrating performance across different model sizes using the \textbf{P}-prompting strategy.}
    \label{fig:distr_deepseek_p}
\end{figure}

As a next step, the two metrics ($F1_{\text{micro}}$~\autoref{sec:metric-f1} and \ac{doc}$_{\text{micro}}$~\autoref{sec:metric-doc}) that make up our overall scoring, let us determine the consistency in accuracy and document level correctness. \autoref{fig:distr_deepseek_p} displays the distribution of results for the \textit{Deepseek-R1}-family. 

We calculated the $F1_{\text{micro}}$-score for every single test case and look at the distribution of results. As the \ac{doc}$_{\text{micro}}$-score is a document level score, it \emph{can not} be calculated for each test, but instead will be calculated separately for each of the eight test scenarios per model testing extend. We observe an steady accumulation of high $F1_{\text{micro}}$-scores, represented by the average low height of blue boxes. This tells us, that model accuracy is fairly consistent between different extraction task and model size variants.

\begin{figure}[!tb]
    \centering
    \includegraphics[width=1\linewidth]{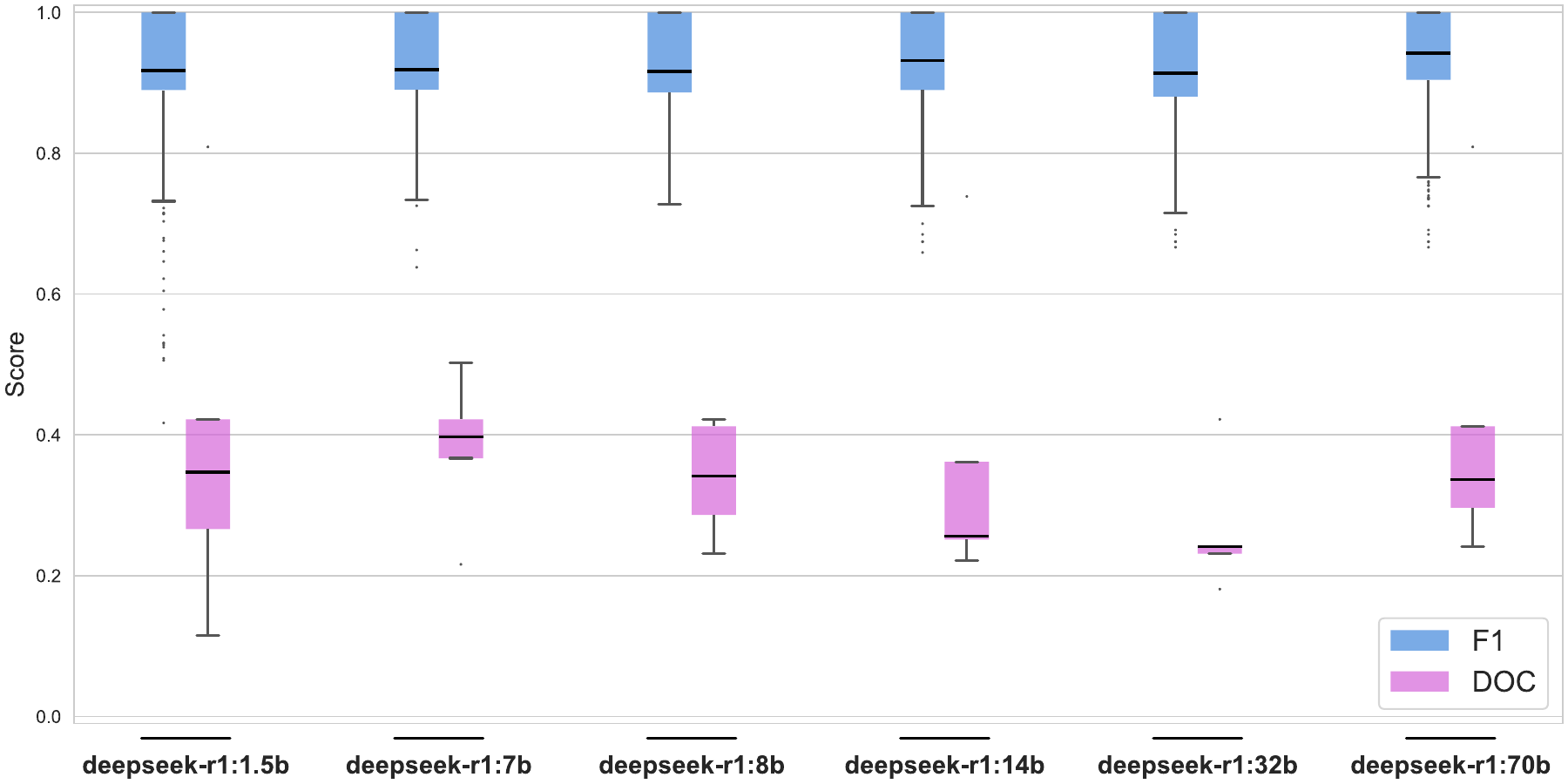}
    \caption{Distribution of $F1_{\text{micro}}$ (blue boxes) and $DOC_{\text{micro}}$ (purple boxes) scores for the \textit{Deepseek-R1} model family, illustrating performance across different model sizes using the \textbf{PJ+}-prompting strategy.}
    \label{fig:distr_deepseek_pj+}
\end{figure}

When we take a look at the \ac{doc}$_{\text{micro}}$-score distribution on the other hand, we do not only observe quite a large variance, but also an increase of variance with a rise in model size. Surprisingly, the same trend is visible with the \textit{Qwen3} and \textit{Gemma3} model-families. In these cases, with larger model size, the median \ac{doc}$_{\text{micro}}$ score rises, but the variance in score is also becoming larger. The cause for this behavior could be down to an incompatibility of prompting strategy with each model family, as we are currently looking at results using \textbf{P}-type prompting. For the \textit{Deepseek-R1} model in specific, there are some jumps in document level correctness variance every other model size. This can be attributed to the two model-families, (\textit{Llama} and \textit{Qwen}), \textit{Deepseek} is derived from, each reacting differently to the applied prompting strategy. To find out if this is the case, we take a look at \autoref{fig:distr_deepseek_pj+} implementing the \textbf{PJ+}-strategy.

Starting with the $F1_{\text{micro}}$-score again, we notice almost no difference, but only minor improvements between the two charts. But comparing the \ac{doc}$_{\text{micro}}$-score reveals, that \textit{Deepseek-R1} does perform way more predictable in terms of correct to failed \ac{json} results with \textbf{PJ+} prompting. Even tough a slight variance in median values between the different model size variants can still be observed, the spread of test scores has generally declined for all model size variants. This confirms, that just changing the prompting strategy, can improve \ac{doc}$_{\text{micro}}$ performance greatly with model families, derived from multiple, different base models.

When we test and score all selected models, it results in the model score list \autoref{tab:model-leaderboard}.

\begin{table}[tbp]
\centering
\caption{Model performance leaderboard displaying $F1_{\text{micro}}$ (token-level accuracy), $DOC_{\text{micro}}$ (document-level validity), and their weighted composite score for all evaluated models using the \textbf{P}-prompting strategy. \textit{GPT-4o} is included as a closed-weight reference.}
\label{tab:model-leaderboard}
\rowcolors*{3}{gray!12}{white}
\setlength{\tabcolsep}{4pt}

\begin{tabularx}{\linewidth}{@{\extracolsep{\fill}}Y c c c c}
\rowcolor{white}
\textbf{Model} & \textbf{Size} \faArrowUp & $F1_{\text{micro}}$ & \ac{doc}$_{\text{micro}}$ & \textbf{Score} \\
\toprule
\textbf{GPT-4o} & $1.8T$* & $0.96$ & $0.52$ & $0.74$ \\
\midrule \addlinespace
\textbf{Gemma3} \faAward     & $27$B  & $\mathbf{0.96}$ & $\mathbf{0.52}$ & $\mathbf{0.74}$ \\
\textbf{Llama3.1}   & $70$B  & $0.95$ & $0.51$ & $0.73$ \\
\textbf{Gemma3}     & $12$B  & $0.95$ & $0.49$ & $0.72$ \\
\textbf{Deepseek-R1} & $70$B  & $0.95$ & $0.45$ & $0.70$ \\
\textbf{Llama3.3}   & $70$B  & $0.95$ & $0.45$ & $0.70$ \\
\textbf{Qwen3}      & $14$B  & $0.95$ & $0.44$ & $0.69$ \\
\textbf{Deepseek-R1} & $14$B  & $0.95$ & $0.43$ & $0.69$ \\
\textbf{Qwen3}      & $1.7$B & $0.94$ & $0.40$ & $0.67$ \\
\textbf{Deepseek-R1} & $7$B   & $0.94$ & $0.39$ & $0.67$ \\
\textbf{Qwen3}      & $32$B  & $0.94$ & $0.39$ & $0.67$ \\
\textbf{Deepseek-R1} & $8$B   & $0.94$ & $0.38$ & $0.66$ \\
\textbf{Phi3}       & $14$B  & $0.94$ & $0.38$ & $0.66$ \\
\textbf{Phi4-mini}  & $3.8$B & $0.95$ & $0.38$ & $0.66$ \\
\textbf{Phi4}       & $14$B  & $0.94$ & $0.38$ & $0.66$ \\
\textbf{Qwen3}      & $4$B   & $0.93$ & $0.37$ & $0.65$ \\
\textbf{Qwen3}      & $8$B   & $0.93$ & $0.36$ & $0.65$ \\
\textbf{Deepseek-R1} & $32$B  & $0.94$ & $0.33$ & $0.63$ \\
\textbf{Qwen3}      & $0.6$B & $0.92$ & $0.30$ & $0.61$ \\
\textbf{Deepseek-R1} & $1.5$B & $0.90$ & $0.16$ & $0.53$ \\
\textbf{Gemma3}     & $1$B   & $0.89$ & $0.10$ & $0.49$ \\
\textbf{Phi3.5}     & $3.8$B & $0.95$ & $0.03$ & $0.49$ \\
\textbf{Phi3}       & $3.8$B & $0.95$ & $0.01$ & $0.48$ \\
\midrule \addlinespace
\rowcolor{white}
\multicolumn{5}{l}{\footnotesize * This is an approximation. No valid information available. } \\
\bottomrule
\end{tabularx}
\end{table}

The small \textit{Phi}-family models generally do not perform good in this evaluation which is down to their incompatibility with the \textbf{P}-prompting strategy as previously noted. We can observe some outliers with \textit{Phi3 - 14b}, \textit{Phi4 - 14b} and \textit{Phi4-mini - 3.8b}, which rank in the middle of our list, implying a better performance with the selected prompting method than their other family members. \\

%


Surprisingly, the leaderboard shows \emph{no strict monotonic relationship} between parameter count and overall score.
While the very top tier is still dominated by large-capacity systems ($\ge{}70$\,B),
the correlation flattens quickly, confirming prior observations that structured-output reliability is often shaped more by prompting and schema handling than by scale alone~\cite{Yaxi_2025,geng2025,chen2024}.

\begin{itemize}
    \item \textbf{Gemma3 – 12b} punches above its weight, landing in the top three and even eclipsing several 70\,B models.
    This mirrors findings from independent structured-output benchmarks, where smaller, well-tuned models achieved accuracy comparable to much larger ones, suggesting that data curation and architectural efficiency can outweigh raw size~\cite{yue2024}.
    \item \textbf{Qwen3 – 1.7b} is only one-tenth the size of the front-runners yet sits comfortably mid-table, outscoring half of the 14–32\,B class. Similar to the conclusions of Lu~\etal~\cite{Yaxi_2025}, who show that model scale alone does not ensure schema-compliant JSON generation, our results highlight the effectiveness of careful prompt alignment even for compact models.
    \item Conversely, \textbf{Deepseek-R1 – 32b} lags behind smaller peers, illustrating that additional parameters alone do not guarantee gains with the chosen \textbf{P}-prompting strategy. This also aligns with prior benchmark analyses where performance differences between decoding frameworks and prompting techniques exceeded those between model sizes~\cite{geng2025}.
\end{itemize}

Taken together, these outliers suggest that pre-training data quality, instruction tuning, and architectural tweaks can outweigh brute-force scale for this particular extraction task.
Comparable effects were reported in synthetic and structured-generation studies, where post-processing design and two-step prompting strategies improved structural validity more than model enlargement~\cite{kashyap2025,li2024}.

\begin{figure}[!tb]
    \centering
    \includegraphics[width=1\linewidth]{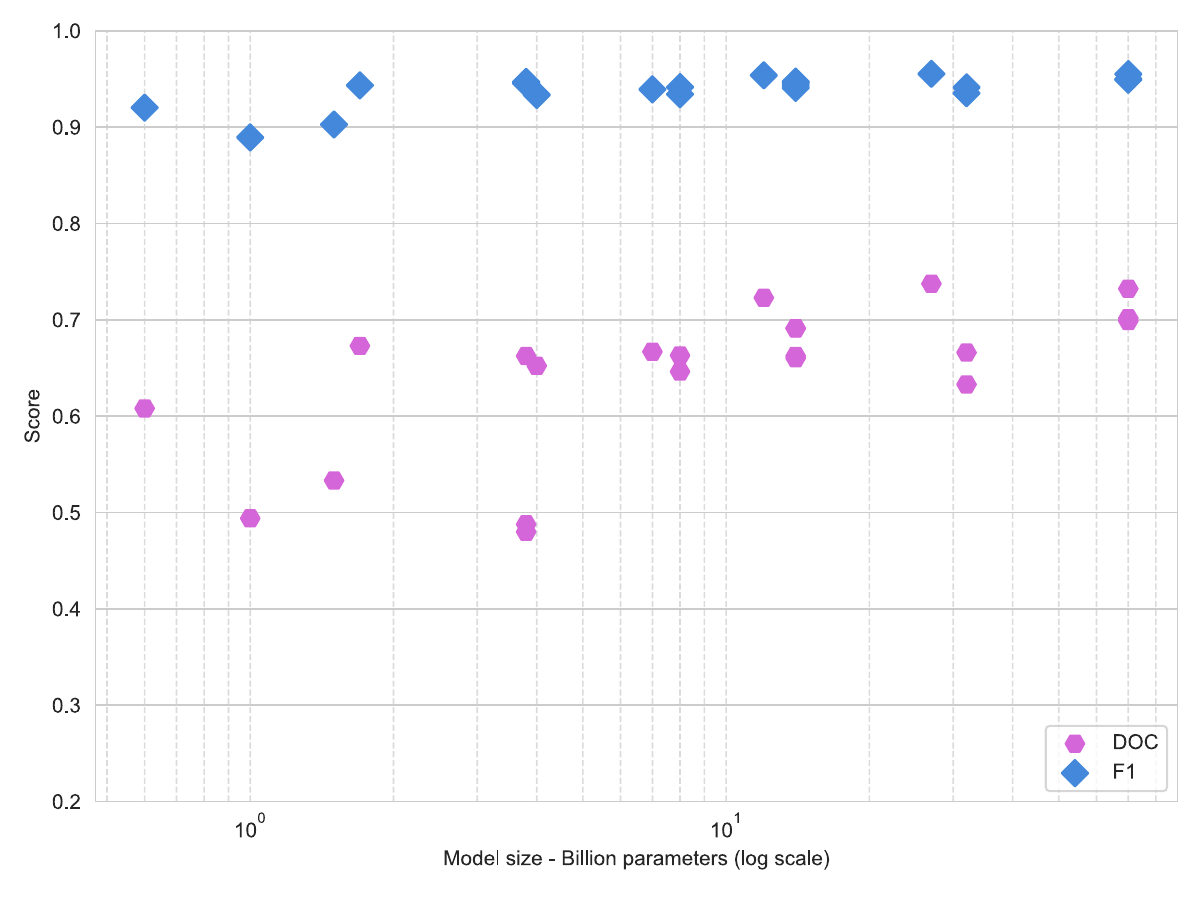}
    \caption{Scatter plot illustrating $F1_{\text{micro}}$ and $DOC_{\text{micro}}$ scores for all models as a function of their logarithmic parameter size, evaluated under the \textbf{P}-prompting strategy.}
    \label{fig:all-scores-p}
\end{figure}

\begin{figure}[!tb]
    \centering
    \includegraphics[width=1\linewidth]{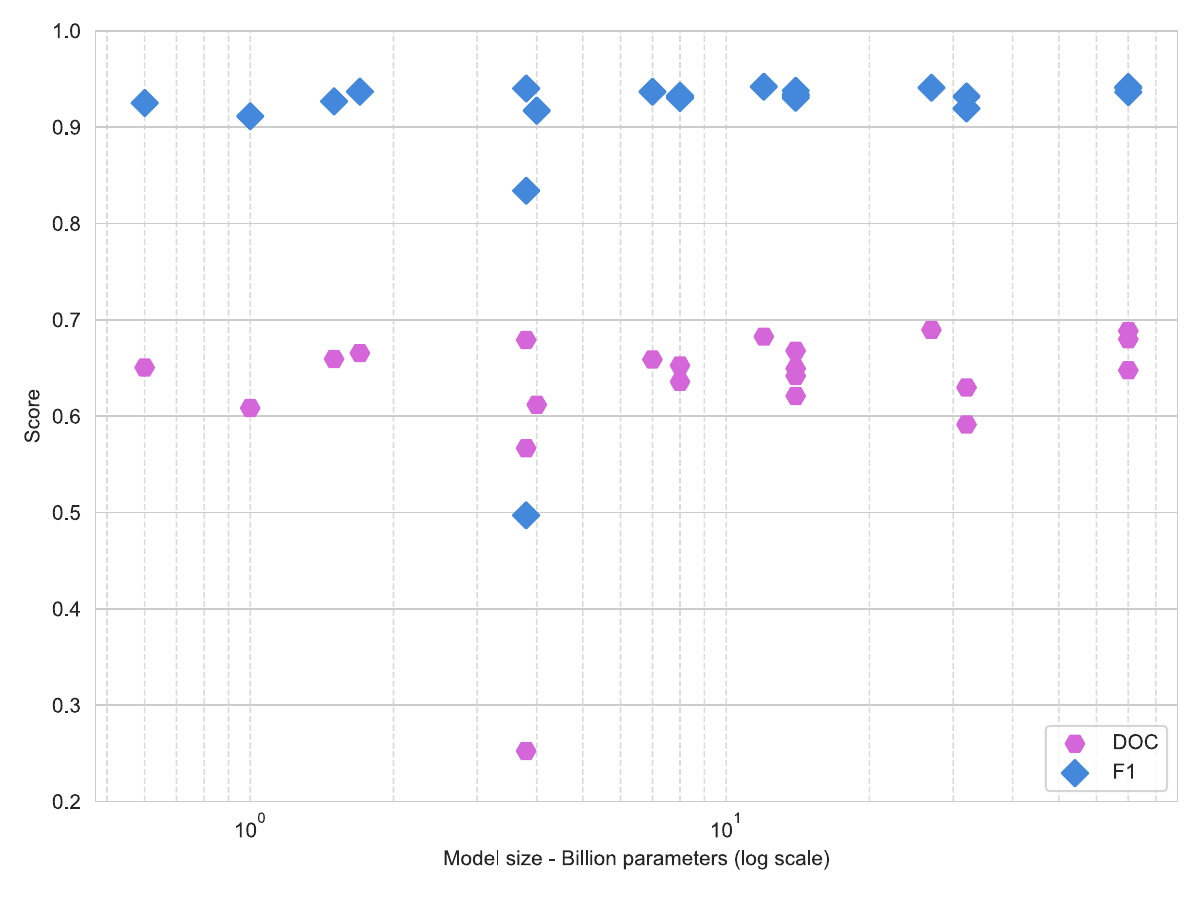}
    \caption{Scatter plot illustrating $F1_{\text{micro}}$ and $DOC_{\text{micro}}$ scores for all models as a function of their logarithmic parameter size, evaluated under the \textbf{PJ+}-prompting strategy.}
    \label{fig:all-scores-pj+}
\end{figure}

The figure shows a steady, largely log-linear improvement in $F1_{\text{micro}}$ as model size increases. In contrast, \ac{doc}$_{\text{micro}}$ exhibits greater variance, particularly among smaller models, and highlights the notable outliers discussed earlier. This suggests that while accuracy generally scales with model size, the document-level score is more sensitive to prompting strategy and model-prompt interaction. To further assess this observation, we examine \autoref{fig:all-scores-pj+}, which presents results for the second-best prompting method, \textbf{PJ+}. \\
 
The $F1_{\text{micro}}$ results closely match the previous prompting setup, with the exception of the Phi models (\textit{Phi3-3.8B} and \textit{Phi3.5-3.8B}), which show reduced accuracy. In contrast, the distribution of \ac{doc}$_{\text{micro}}$ scores shifts notably. No clear monotonic trend appears across model sizes, and the greatest variation now occurs in mid-sized models rather than at the small end. This further supports our finding that accuracy scales primarily with model size, whereas document-level reliability depends strongly on the combination of model and prompting strategy rather than scale alone.

\subsection{Detailed Failure Analysis}\label{sec:leaderboard-results} 
\begin{figure*}[!t]
    \centering
    \includegraphics[width=1\linewidth]{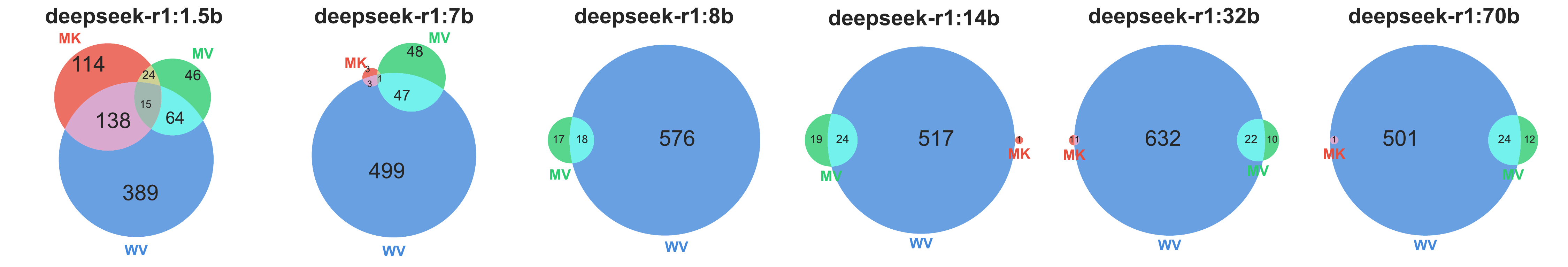}
    \caption{Venn diagrams illustrating the distribution and overlap of specific error types (\acf{mk}, \acf{wv}, \acf{mv}) for the \textit{Deepseek-R1} model family under the \textbf{P}-prompting strategy. Each segment indicates the count of test cases exhibiting a single error type or a combination of types.}
    \label{fig:venn-p-deepseek}
\end{figure*}

Another interesting point of investigation is the distribution of \ac{json} error types per count of performed tests. This will give us an insight, on what kind errors were made and on how many tests a specific combination of errors occurred. To visualize the overlapping distributions, we opted to imply a \textit{venn} plot. On \autoref{fig:venn-p-deepseek}, the error types \textit{\ac{mk}}, \textit{\ac{wv}}, \textit{\ac{mv}} are shown for the \textit{Deepseek-R1} model family, thrown using the \textbf{P}-prompting strategy. An overlap of two bubbles represents, that both mistakes were made at least once in the same test. As only one count of error category is counted per single test, the combined value of all bubbles in one plot is equal to the count of all tests ($995$) minus the individual failed and flawless results per model. For each checkpoint the count of runs that could exhibit any error are calculated as follows.
\[
N_{\text{err}} = N_{\text{tests}}-\bigl(N_{\text{perfect}}+N_{\text{failed}}\bigr),
\]

The \textit{Deepseek-R1} series shows a clear trajectory under the \textbf{P}-prompting setup. At \textit{1.5b} parameters the model still drops required keys frequently, making schema violations a noticeable portion of its failures. That problem is largely resolved by the time the network reaches \textit{7b}, where only three such cases remain, and it effectively disappears at \textit{8b} and beyond, with at most a single omission recorded for any larger checkpoint. \\

Once keys are reliably present, the dominant issue becomes incorrect content. \ac{wv} errors start at $389$ cases for \textit{1.5b}, climb steadily with scale and peak at $632$ for the \textit{32b} checkpoint before edging down to $501$ at \textit{70b}. Even the largest model therefore produces at least one wrong field in roughly half of the test runs, indicating that \emph{value fidelity, rather than structural compliance}, is the main bottleneck.

Purely \ac{mv} slightly decline from $46$ at \textit{1.5b} to $12$ at \textit{70b}, tracking a similar drop in the overlapping \textit{missing \& wrong} region. The overlap never vanishes entirely, implying that the two error modes share a common source of confusion about the required content and structure. \\

In summary, key omissions vanish once the model reaches mid-scale, but incorrect or absent values persist. Larger parameter counts beyond roughly \textit{14b} offer only marginal gains for value accuracy, so \emph{further improvement is likely to depend on specialized conditioning or post-generation validation rather than additional scale alone}.\\

\begin{figure*}[!t]
    \centering
    \includegraphics[width=1\linewidth]{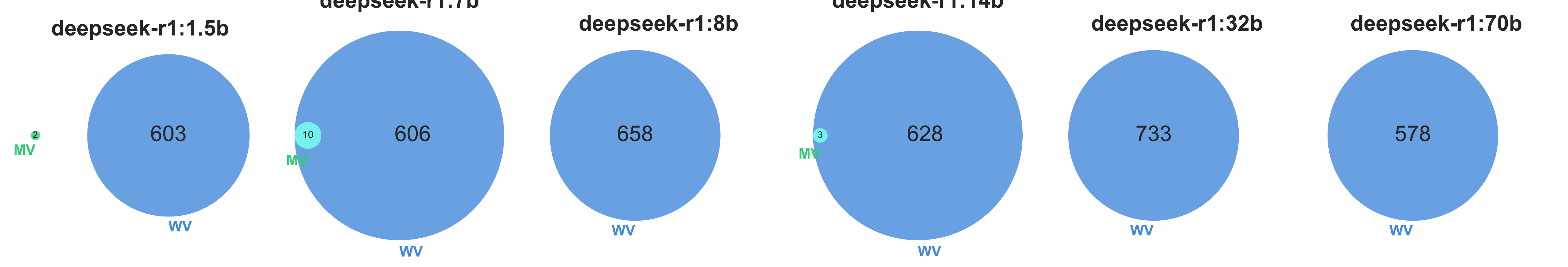}
    \caption{Venn diagrams illustrating the distribution and overlap of specific error types (\acf{mk}, \acf{wv}, \acf{mv}) for the \textit{Deepseek-R1} model family under the \textbf{PJ+}-prompting strategy. Each segment indicates the count of test cases exhibiting a single error type or a combination of types.} 
    \label{fig:venn-pj+-deepseek}
\end{figure*}

Switching from the \textbf{P} to the \textbf{PJ+} template on \autoref{fig:venn-pj+-deepseek} changes the error \emph{mix} more than the error \emph{volume}. To better understand the differences in error spread, we consult the \autoref{tab:err-dist} that shows \(N_{\text{err}}\) for each model and the two selected prompting options combined with inclusive incidence of the three error classes. Individual error shares are calculated relative to \(N_{\text{err}}\). As there is overlap of error types in between a single tests, the sum of all percentages can exceed \(100\,\%\).

\noindent

\begin{table}[tbp]
\centering
\caption{Absolute number of erroneous test cases (\(N_{\text{err}}\)) and the percentage incidence of \ac{mk}, \ac{mv}, and \ac{wv} errors for the \textit{Deepseek-R1} model family, across both \textbf{P} and \textbf{PJ+} prompting strategies. Percentages are calculated relative to \(N_{\text{err}}\) and can sum to more than 100\% due to overlapping error types.}
\label{tab:err-dist}
\begin{tabularx}{\linewidth}{@{\extracolsep{\fill}}l cccc | cccc}
\midrule
 & \multicolumn{4}{c}{\textbf{P}} & \multicolumn{4}{c}{\textbf{PJ+}}\\
\cmidrule(lr){2-5}\cmidrule(l){6-9}
 & & \multicolumn{3}{c}{\tiny\(\%\)} & & \multicolumn{3}{c}{\tiny\(\%\)}\\
 \cmidrule(lr){3-5}\cmidrule(l){7-9}
Model & \(N_{\text{err}}\) & \ac{mk} & \ac{mv} & \ac{wv} & \(N_{\text{err}}\) & \ac{mk} & \ac{mv} & \ac{wv}\\
\midrule
$1.5$ B & $790$ & $36.8$ & $18.9$ & $76.7$ & $605$ & $0.0$ & $0.3$ & $99.7$\\
$7$ B   & $601$ &  $1.2$ & $16.0$ & $91.5$ & $616$ & $0.0$ & $1.6$ & $100.0$\\
$8$ B   & $611$ &  $0.0$ &  $5.7$ & $97.2$ & $658$ & $0.0$ & $0.0$ & $100.0$\\
$14$ B  & $561$ &  $0.2$ &  $7.7$ & $96.4$ & $631$ & $0.0$ & $0.5$ & $100.0$\\
$32$ B  & $666$ &  $0.3$ &  $4.8$ & $98.3$ & $733$ & $0.0$ & $0.0$ & $100.0$\\
$70$ B  & $538$ &  $0.2$ & $6.7$ & $97.8$ & $578$ & $0.0$ & $0.0$ & $100.0$\\
\bottomrule
\end{tabularx}
\end{table}

\begin{table}[tbp]
\centering
\setlength{\tabcolsep}{3pt}

\caption{Absolute number of erroneous test cases (\(N_{\text{err}}\)) and the percentage incidence of \ac{mk}, \ac{mv}, and \ac{wv} errors for the top-performing models (\textit{GPT-4o} and \textit{Gemma3 - 27b}) across both \textbf{P} and \textbf{PJ+} prompting strategies. Percentages are calculated relative to \(N_{\text{err}}\) and can sum to more than 100\% due to overlapping error types.}
\label{tab:err-dist-top}
\begin{tabularx}{\linewidth}{@{\extracolsep{\fill}}lcccc | cccc@{}}
\midrule
 & \multicolumn{4}{c}{\textbf{P}} & \multicolumn{4}{c}{\textbf{PJ+}}\\
\cmidrule(lr){2-5}\cmidrule(l){6-9}
 & & \multicolumn{3}{c}{\tiny\(\%\)} & & \multicolumn{3}{c}{\tiny\(\%\)}\\
 \cmidrule(lr){3-5}\cmidrule(l){7-9}
Model & \(N_{\text{err}}\) & \ac{mk} & \ac{mv} & \ac{wv} & \(N_{\text{err}}\) & \ac{mk} & \ac{mv} & \ac{wv}\\
\midrule
\textbf{GPT-4o} & $473$ & $0.0$ & $9.3$ & $96.4$ & $555$ & $0.0$ & $5.0$ & $98.0$ \\
\textbf{Gemma3 - 27b} & $478$ & $0.0$ & $6.7$ & $97.3$ & $559$ & $0.0$ & $0.0$ & $100.0$ \\
\bottomrule
\end{tabularx}

\end{table}

\noindent
Under the \textbf{P} prompt, \ac{mk} errors still account for \(36.8\,\%\) of the error pool at \textit{1.5b} but drop below \(1.2\,\%\) by \textit{7b}, while \ac{mv} persists between \(4.8,\%\) and \(18.9\,\%\) across the model size range. Inclusive counting shows that \ac{wv} already describes most imperfect runs, reaching \(97\text{–}98\,\%\) on the larger checkpoints. \\

For the \textit{1.5b} checkpoint, replacing \textbf{P} with \textbf{PJ+} lowers \(N_{\text{err}}\) from \(790\) to \(605\) (\(-23\,\%\)), because many outputs that previously failed the schema test now pass.
For all larger checkpoints \(N_{\text{err}}\) rises modestly by $2$ to $13$ per-cent. Since \textbf{PJ+} template leaves \emph{no} run unparseable yet converts some formerly perfect parses into \ac{wv} cases (e.g.\ \(433\) perfect runs at \textit{14b} under \textbf{P} shrink to \(364\) under \textbf{PJ+}).\\

In practical terms, \textbf{PJ+} guarantees a complete \ac{json} structure with the \textit{Deepseek-R1} model family but redirects almost every remaining uncertainty into field content. The result is a higher count of semantically flawed values despite the absence of structural faults, \emph{shifting the downstream burden from schema repair to semantic validation.}

A comparison with the top-performing closed- and open-weight systems is shown in \autoref{tab:err-dist-top}, where \textit{GPT-4o} and \textit{Gemma3 - 27b} exhibit a similar dominance of \ac{wv} errors under both prompting strategies.

\section{Conclusion}\label{sec:conclusions}
\noindent
While \textit{GPT-4o} was added as a closed-weight reference, it did not exhibit a clear advantage over the best open-source models (e.g. \textit{Gemma3 - 27B}), reinforcing the practical competitiveness of deployable open systems.

The experiments further prove that prompting strategy and model architecture are at least as important as model scale for \ac{json} parsing tasks.
\begin{itemize}
    \item \textbf{P} strategy often yields the best overall balance of accuracy and validity for well-aligned models but can fail catastrophically for some families (e.g., \textit{Phi}).
    \item \textbf{PJ+} strategy is the safest choice for ensuring parseable outputs, especially for small or structurally unreliable models, but this comes at the cost of increased semantic errors.
    \item Model size helps, particularly for improving value accuracy $\bigl(F1_{\text{micro}}\bigr)$, but \ac{doc}$_{\text{micro}}$ is more sensitive to prompt–model compatibility than to scale.
    \item \ac{wv} errors remain the primary bottleneck across all configurations, suggesting that post-generation semantic validation or fine-tuning is needed for further gains.
    \item Optimal performance is achieved by matching the prompting approach to the model family, not by relying solely on large parameter counts. \\
\end{itemize}
As shown in \autoref{tab:model-performance-summary}, we summarize the best-performing prompting strategy for each evaluated model family and size, along with their composite scores under both the \textbf{P} and \textbf{PJ+} prompting strategy, the dominant error types, and notable observations. This overview condenses the experimental findings by linking model scale, architecture, and prompt design to observed strengths and weaknesses. It highlights cases where smaller models achieve competitive performance, where prompting choice has a decisive impact on document validity $\bigl(DOC_{\text{micro}}\bigr)$, and where specific error types, such as \ac{wv}, persist regardless of structural correctness.

\begin{table*}[tbp]
\centering
\caption{Comprehensive summary of model performance, showing the best-performing prompting strategy for each model family and size variant. The table includes composite scores under both \textbf{P} and \textbf{PJ+} prompting strategy, identifies dominant error types, and provides key observations regarding model-specific strengths and weaknesses.}
\label{tab:model-performance-summary}
\begin{tabular*}{\linewidth}{@{\extracolsep{\fill}}ll c ll ll}
\midrule
\textbf{Model} & \textbf{Size} & \textbf{Strategy} & \textbf{Score (P)} & \textbf{Score (PJ+)} & \textbf{Dominant Error(s)} & \textbf{Notable Observations} \\
\midrule
Gemma3 & 1B   & PJ+ & 0.49 & $\uparrow_{DOC_{\text{micro}}} +0.2$  & \ac{mk}, \ac{mv} & Complexity penalty under \textbf{P}; \textbf{PJ+} improves structure \\
       & 12B  & P   & 0.72 & $\approx$ & \ac{wv} & Outperforms some \textit{70B} models; big jump from \textit{1B} \\
       & 27B  & P   & 0.74 & $\approx$ & \ac{wv} & Marginal gain over \textit{12B}; structural errors rare \\
\midrule
LLaMA3 & 70B (3.1/3.3) & P & 0.70--0.73 & $\approx$ & \ac{wv} & High accuracy; $DOC_{\text{micro}}$ variance increases with size \\
\midrule
Qwen3  & 0.6B & PJ+ & 0.61 & $\uparrow$ & \ac{mk}, \ac{wv} & Low $DOC_{\text{micro}}$ with \textbf{P}; \textbf{PJ+} removes structural errors \\
       & 1.7B & P   & 0.67 & $\approx$ & \ac{wv} & Mid-table despite small size \\
       & 4--32B & P & 0.65--0.69 & $\approx$ & \ac{wv} & \textit{32B} no better than smaller peers \\
\midrule
Deepseek-R1 & 1.5B & PJ+ & 0.53 & $\uparrow_{N_{\text{err}}}$ $-23\%\,$ & \ac{mk}, \ac{mv}, \ac{wv} & \textbf{PJ+} eliminates \ac{mk}; shifts errors to \ac{wv} \\
            & 7--14B & P & 0.67--0.69 & $\approx$ & \ac{wv} & \ac{mk} nearly absent; \textbf{PJ+} slightly increases \ac{wv} rate \\
            & 32B & P & 0.63 & $\downarrow$ & \ac{wv} & Underperforms smaller peers \\
            & 70B & P & 0.70 & $\approx$ & \ac{wv} & High $F1_{\text{micro}}$ but persistent \ac{wv} issues \\
\midrule
Phi    & 3.8B (Phi3/3.5) & PJ+ & 0.48--0.49 & $\downarrow$ or $\approx$ & \ac{wv} & \textbf{P} poor; \textbf{PJ+} mitigates structural failures \\
       & 14B (Phi3/4)    & P   & 0.66 & $\approx$ & \ac{wv} & Mid-table despite family trend \\
       & 3.8B (Phi4-mini)& P   & 0.66 & $\approx$ & \ac{wv} & Outlier: better than expected \\
\bottomrule
\end{tabular*}
\end{table*}


\noindent 

\newpage
\bibliographystyle{IEEEtran}
\bibliography{IEEEabrv,sources}
\appendix

\subsection{Formal definition of the micro-F1 metric}
\label{app:microf1_formula}

For completeness, we restate the exact formulation of the micro-averaged score $F1_{\text{micro}}$ used throughout this work.
\begin{align}
F1_{\text{micro}}
&= \alpha\,F1_{\text{keys}} + (1-\alpha)\,F1_{\text{values}}, \qquad \alpha = 0.25,
\end{align}
where each component is defined as a standard $F_1$–measure:
\begin{align}
\label{eq:f1_pr_defs} 
F1_{\text{keys}} &= \frac{2\,P_{\text{keys}}\,R_{\text{keys}}}
     {P_{\text{keys}} + R_{\text{keys}}}, & 
F1_{\text{values}} &= \frac{2\,P_{\text{values}}\,R_{\text{values}}}
     {P_{\text{values}} + R_{\text{values}}}, \\[8pt] 
P_{\text{keys}} &= \frac{TP_k}{TP_k + FP_k}, & 
P_{\text{values}} &= \frac{TP_v}{TP_v + FP_v}, \\[8pt] 
R_{\text{keys}} &= \frac{TP_k}{TP_k + FN_k}, & 
R_{\text{values}} &= \frac{TP_v}{TP_v + FN_v}.
\end{align}

\noindent
The value-level true positives $TP_v$ incorporate partial credit for near matches:
\begin{align}
\label{eq:tpv_broken}
TP_v &= N_v - FN_v - (E_{\text{type}} + E_{\text{value}} + E_{\text{coerce}} + E_{\text{dev}}) \\ 
     &\quad + \sum_{i=1}^{N_{\text{lev}}}C(l_i) + \beta\,E_{\text{coerce}}, \notag 
\end{align}
with $\beta=0.2$ denoting the similarity gain for \emph{coercible mismatches}.
The fuzzy credit function $C(l)$ depends on the normalized Levenshtein distance $l$ between reference and prediction:
\begin{align}
C(l) =
\begin{cases}
1, & l \le L_{\text{good}},\\[2pt]
\gamma\,(1-l), & L_{\text{good}} < l < L_{\text{bad}},\\[2pt]
p_{\text{bad}}, & l \ge L_{\text{bad}},
\end{cases}
\end{align}
where $\gamma = 0.5$ controls the weight for approximate matches,
$L_{\text{good}} = 0.1$, $L_{\text{bad}} = 2.0$, and $p_{\text{bad}} = -1$ applies a penalty for hopelessly wrong strings.

\definecolor{jsonbg}{HTML}{F8F9FB}
\definecolor{jsonkey}{HTML}{1F6FEB}
\definecolor{jsonstring}{HTML}{0A7C2F}
\definecolor{jsonnumber}{HTML}{B54800}
\definecolor{jsonbool}{HTML}{A626A4}
\definecolor{jsonnull}{HTML}{6E7781}
\definecolor{jsonpunct}{HTML}{6E7781}
\definecolor{jsonframe}{HTML}{E5E7EB}

\lstdefinelanguage{json}{
  sensitive=true,
  showstringspaces=false,
  showtabs=false,
  morestring=[b]",
  stringstyle=\color{jsonstring},
  morecomment=[l]{//},
  morecomment=[s]{/*}{*/},
  commentstyle=\itshape\color{gray},
  literate=
   *{:}{{{\color{jsonpunct}{:}}}}{1}
    {,}{{{\color{jsonpunct}{,}}}}{1}
    {\{}{{{\color{jsonpunct}{\{}}}}{1}
    {\}}{{{\color{jsonpunct}{\}}}}}{1}
    {[}{{{\color{jsonpunct}{[}}}}{1}
    {]}{{{\color{jsonpunct}{]}}}}{1}
    {true}{{{\color{jsonbool}{true}}}}{4}
    {false}{{{\color{jsonbool}{false}}}}{5}
    {null}{{{\color{jsonnull}{null}}}}{4}
    {0}{{{\color{jsonnumber}0}}}{1}
    {1}{{{\color{jsonnumber}1}}}{1}
    {2}{{{\color{jsonnumber}2}}}{1}
    {3}{{{\color{jsonnumber}3}}}{1}
    {4}{{{\color{jsonnumber}4}}}{1}
    {5}{{{\color{jsonnumber}5}}}{1}
    {6}{{{\color{jsonnumber}6}}}{1}
    {7}{{{\color{jsonnumber}7}}}{1}
    {8}{{{\color{jsonnumber}8}}}{1}
    {9}{{{\color{jsonnumber}9}}}{1}
    {.}{{{\color{jsonnumber}.}}}{1}
    {-}{{{\color{jsonnumber}-}}}{1},
}

\subsection{Dataset examples}
\label{app:dataset_examples}
To illustrate the diversity and complexity of the \emph{LLMStructBench} dataset, we present examples from two distinct use cases: an easy task, \textit{Support Tickets}, and a harder task, \textit{Loan Request}. The examples showcase the natural language messages and their corresponding \ac{gt} \ac{json} objects.

The \textbf{Support Ticket} use case involves extracting essential information from IT support requests. This task shows a flat key-value structure and relatively straightforward data types.

\begin{lstlisting}[caption={Example Natural Language Message for Support Ticket Task},label={lst:support_ticket_nl_msg}]
Dear IT Support Team,

My name is Aiko Tanaka, and my Employee ID is XY729B. I am writing to report an issue that I am experiencing. I am currently unable to access the shared drives due to a permission error. The severity of this issue is high as it is impacting my ability to work.

This issue was first noted on 05.10.2023. I would greatly appreciate it if you could look into this matter at your earliest convenience.

Thank you for your prompt attention to this issue.

Best regards,
Aiko Tanaka
\end{lstlisting}

\begin{lstlisting}[language=json,caption={Ground Truth JSON for Support Ticket Task},label={lst:support_ticket_gt_json}]
{
    "EmployeeName": "Aiko Tanaka",
    "EmployeeID": "XY729B",
    "IssueDescription": "Unable to access shared drives due to permission error.",
    "IssueSeverity": "High",
    "ReportDate": "05.10.2023"
}
\end{lstlisting}

The \ac{json} schema definition is as follows:
\begin{lstlisting}[language=json,caption={Example JSON schema definition for Support Ticket Task},label={lst:support_ticket_schema}]
{
  "title": "it_support_ticket",
  "description": "Schema defining the structure for reporting an IT support issue, including employee details, issue description, severity, and report date.",
  "type": "object",
  "properties": {
    "EmployeeName": {
      "type": "string",
      "description": "The full name of the employee reporting the issue."
    },
    "EmployeeID": {
      "type": "string",
      "description": "The unique identifier assigned to the employee."
    },
    "IssueDescription": {
      "type": "string",
      "description": "A detailed description of the IT issue being experienced."
    },
    "IssueSeverity": {
      "type": "string",
      "description": "The perceived severity level of the issue (e.g., High, Medium, Low)."
    },
    "ReportDate": {
      "type": "string",
      "description": "The date the issue was reported (DD.MM.YYYY format).",
      "pattern": "^[0-9]{2}\\.[0-9]{2}\\.[0-9]{4}$"
    }
  }
}
\end{lstlisting}

The \textbf{Loan Request} use case requires extracting details about equipment loans, including an array of items with nested properties.

\begin{lstlisting}[caption={Example Natural Language Message for Loan Request Task},label={lst:loan_request_nl_msg}]
Dear Hiroshi Yamamoto,

This is a reminder that the following equipment is due for return on 23.11.2023:

- 1 tablet
- 2 VR headsets
- 1 digital camera

Please ensure all items are returned on or before the due date to avoid any late fees or penalties.

Thank you,
[Your Organization's Name]
\end{lstlisting}

\begin{lstlisting}[language=json,caption={Ground Truth JSON for Loan Request Task},label={lst:loan_request_gt_json}]
{
    "BorrowerName": "Hiroshi Yamamoto",
    "BorrowerID": null,
    "Equipment": [
      {
        "EquipmentType": "tablet",
        "EquipmentQuantity": 1
      },
      {
        "EquipmentType": "VR headset",
        "EquipmentQuantity": 2
      },
      {
        "EquipmentType": "digital camera",
        "EquipmentQuantity": 1
      }
    ],
    "ReturnDate": "23.11.2023"
}
\end{lstlisting}

The \ac{json} schema definition is as follows:
\begin{lstlisting}[language=json,caption={Example JSON schema definition for Loan Request Task},label={lst:loan_request_schema}]
{
  "title": "equipment_loan_request",
  "description": "Schema defining the structure for an equipment borrowing record, including borrower details, borrowed items, and the expected return date.",
  "type": "object",
  "properties": {
    "BorrowerName": {
      "type": "string",
      "description": "The full name of the person borrowing the equipment."
    },
    "BorrowerID": {
      "type": "string",
      "description": "A unique identifier for the borrower."
    },
    "Equipment": {
      "type": "array",
      "description": "A list of equipment items being borrowed.",
      "items": {
        "type": "object",
        "description": "Details of a specific type of equipment being borrowed.",
        "properties": {
          "EquipmentType": {
            "type": "string",
            "description": "The type or name of the equipment (e.g., 'tablet', 'VR headset')."
          },
          "EquipmentQuantity": {
            "type": "integer",
            "description": "The number of units of this equipment type being borrowed.",
            "minimum": 1
          }
        },
        "required": [
          "EquipmentType",
          "EquipmentQuantity"
        ],
        "additionalProperties": false
      }
    },
    "ReturnDate": {
      "type": "string",
      "description": "The date when the equipment is expected to be returned (DD.MM.YYYY format).",
      "pattern": "^[0-9]{2}\\.[0-9]{2}\\.[0-9]{4}$"
    }
  }
}
\end{lstlisting}


 




\vfill 

\end{document}